\documentclass[APA,LATO1COL]{WileyNJD-v2}

\usepackage{amsmath}
\usepackage{graphicx}
\usepackage{subfig}
\usepackage{url}
\usepackage{hyperref}

\usepackage{amsmath,amssymb,amsfonts}
\usepackage{textcomp}
\usepackage{multirow}
\usepackage{multicol}
\usepackage{color}
\usepackage{caption}

\usepackage{booktabs}
\usepackage{tikz}
\usetikzlibrary{arrows,positioning}
\usepackage{lipsum, adjustbox}

\usepackage{setspace}

\articletype{Original Article}%
\DeclareMathOperator*{\argmin}{arg\,min}

\newcommand{\setItemSep}{\setlength\itemsep}

\newcommand{\setFeature}{\mathcal{S}}
\newcommand{\setFeatureVal}{s} 
\newcommand{\numFeatsRemovedInc}{k} 
\newcommand{\numFeatsRemoved}{K}
\newcommand{\numFeatsTotal}{J}

\newcommand{\indexFeat}{j}
\newcommand{\indexSlice}{\ell}

\newcommand{\instance}{x}
\newcommand{\instanceVect}{\mathbf{\instance}}
\newcommand{\instanceMat}{\mathbf{X}}

\newcommand{\windowSize}{L}

\newcommand{\impScore}{\phi}
\newcommand{\impScoreMat}{\Phi}

\newcommand{\neuralNetWeights}{\Theta}

\newcommand{\timeVarTotal}{T}
\newcommand{\timeHorizon}{\tau}
\newcommand{\timeHorizonTotal}{t_0}

\newcommand{\expectedVal}{\mathbb{E}_r}

\newcommand{\aptThreshold}{\alpha}

\newcommand{\tcolB}{\textcolor{blue}}

\begin{document}

\title{Evaluation of Local Explanation Methods for Multivariate Time Series Forecasting}

\author[1]{Ozan Ozyegen}

\author[1]{Igor Ilic}

\author[1]{Mucahit Cevik}

\authormark{Ozyegen \textsc{et al}}

\address[1]{\orgdiv{Data Science Lab, Mechanical Industrial Engineering Department}, \orgname{Ryerson University}, \orgaddress{\state{Ontario}, \country{Canada}}}

\corres{*Ozan Ozyegen, Data Science Lab, Mechanical Industrial Engineering Department, Ryerson University, Toronto, Ontario, M5B 2K3, Canada. \email{oozyegen@ryerson.ca}}

\abstract[Abstract]{Being able to interpret a machine learning model is a crucial task in many applications of machine learning. Specifically, local interpretability is important in determining why a model makes particular predictions. Despite the recent focus on AI interpretability, there has been a lack of research in local interpretability methods for time series forecasting while the few interpretable methods that exist mainly focus on time series classification tasks. In this study, we propose two novel evaluation metrics for time series forecasting: Area Over the Perturbation Curve for Regression and Ablation Percentage Threshold. These two metrics can measure the local fidelity of local explanation models. We extend the theoretical foundation to collect experimental results on two popular datasets, \textit{Rossmann sales} and \textit{electricity}. Both metrics enable a comprehensive comparison of numerous local explanation models and find which metrics are more sensitive. Lastly, we provide heuristical reasoning for this analysis.}

\maketitle

\section{Introduction}
As machine learning approaches find more use cases in the society, the machine learning systems become more complex and less interpretable. Often times, models are assessed by their prediction performance (e.g. by mean squared error and mean absolute error for regression tasks) on a test set which does not consider the interpretability of the underlying model. In the absence of understanding why a model is making a decision, trusting a model can lead to inaccurate and potentially dangerous decisions~\citep{caruana2017intelligible}. However, in recent years, the value of interpretability in machine learning has been recognized and gained significant traction~\citep{adadi2018peeking, doshi2017towards, guidotti2018survey}.


Higher interpretability has many benefits. First of all, it can create trust by showing the different factors contributing to the decisions~\citep{gilpin2018explaining}. Trust on the model in turn can lead to a higher acceptance of machine learning systems~\citep{adadi2018peeking}. Secondly, interpretability tools can reveal incompleteness in the problem formalization~\citep{doshi2017towards}. This information can then be used to debug the model and design better models. Finally, these methods can be used to improve our scientific understanding~\citep{doshi2017towards}. By analyzing how machine learning models behave, we can enhance knowledge about a topic~\citep{boshra2019neurophysiological}.

Interpretability aims to better understand an automated model. Based on the scope of interpretability we can distinguish the existing methods to two classes: global and local. \textit{Global interpretability} methods try to explain the entire logic and reasoning of a model. On the other hand, \textit{local interpretability} methods aim to explain the reasons for a specific decision~\citep{adadi2018peeking}.

\begin{figure}[!ht]
    \centering
    \includegraphics[width=0.5\columnwidth]{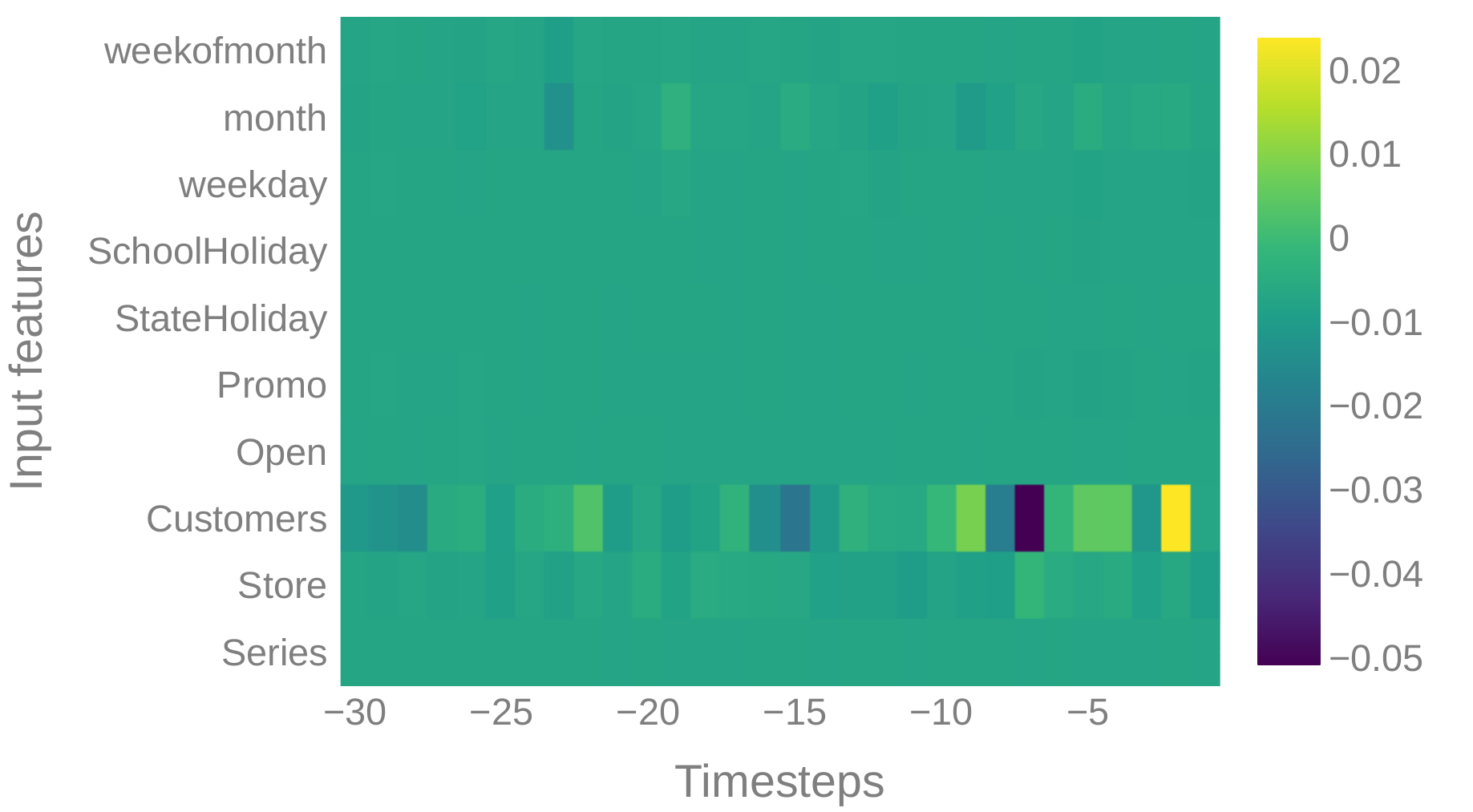}
    \caption{Local explanation of a random sample from the Rossmann dataset obtained from feature importances, where the positively and negatively contributing features are highlighted in yellow and blue, respectively. This sample is indicating that number of customers is the most important feature for the local predictions.}
    \label{fig:heatmap}
\end{figure}

In this paper we focus on local explanations for multivariate time series forecasting, where \textit{multivariate} refers to multiple input features. For a selected sample and forecasting horizon, a local explanation method can be used to show the importance of each input feature to the prediction. Local explanations are two-sided which means they show both the magnitude and the directional importance of the features. A heatmap of feature importances for an arbitrary sample in \textit{Rossmann sales} dataset is shown in Figure~\ref{fig:heatmap} where the positively and negatively contributing features are highlighted in yellow and blue, respectively. Rossmann sales dataset shows the historical store sales of more than a thousand drug stores. All samples from the dataset contain a 30 day time window ($x$-axis) of 10 different features ($y$-axis). By analyzing local explanations over the samples, we can observe how important different features are to the model and how they are contributing to the prediction. For instance, by analyzing Figure~\ref{fig:heatmap}, we can infer that the number of customers is the most important feature for the local predictions (i.e. for the given sample). These feature importances can also be averaged over many samples to compute the global importance of each feature.

Evaluation of local explanations is challenging but necessary to minimize misleading explanations. Various approaches have been used to evaluate local explanations, from visual inspection~\citep{ding2017visualizing} to measuring the impact of deleting important features on the classifier output~\citep{li2016understanding, nguyen2018comparing, olah2020zoom}. However, in time series domain, there are only a few studies focusing on interpretability of the machine learning models~\citep{norgeot2018time, suresh2017clinical}. Moreover, the existing literature is mostly focused on time series classification task~\citep{felix2019shap}. Interpreting time series forecasting is equally important as time series classification, with many applications in various areas such as physics simulations~\citep{rubanova2019latent}, spectrum occupancy prediction~\citep{ozyegen2019experimental} and sales forecasting~\citep{jain2015sales}. However, the research on interpretability of time series forecasting models focus on intrinsic explainability~\citep{lim2019temporal}, and the absence of proper evaluation measures for local explainability methods for time series forecasting tasks might be considered as one possible reason for the relatively low research interest on interpreting time series forecasting models. That is, a better understanding on properly evaluating local explanations would potentially contribute to a further progress in the area. Accordingly, our paper makes the following contributions:
\begin{itemize}\setItemSep{0.3em}
    \item \textit{Novel evaluation metrics for time series forecasting}: We introduce two novel evaluation metrics for comparing local interpretability methods which can be applied on any type of time series forecasting problem.
    
    \item \textit{Comparison of local explanation methods for multivariate time series forecasting}: We perform a comprehensive comparison of three local explanation approaches (and a random baseline) on two different datasets.
\end{itemize}


\section{Related Work}

Machine learning is used to improve many products and processes. On the other hand, a large barrier for adopting machine learning in many systems is the black-box architecture of many of the machine learning systems~\citep{molnar2019interpretable}. 


Interpretable AI can be considered as a toolbox which consists of many different methods. While different taxonomies were proposed, we focus on the one proposed by~\citet{adadi2018peeking} where the existing interpretable AI approaches are classified under three criteria: complexity, scope and model related.

\begin{figure}[!ht]
    \centering
    \includegraphics[width=0.5\columnwidth]{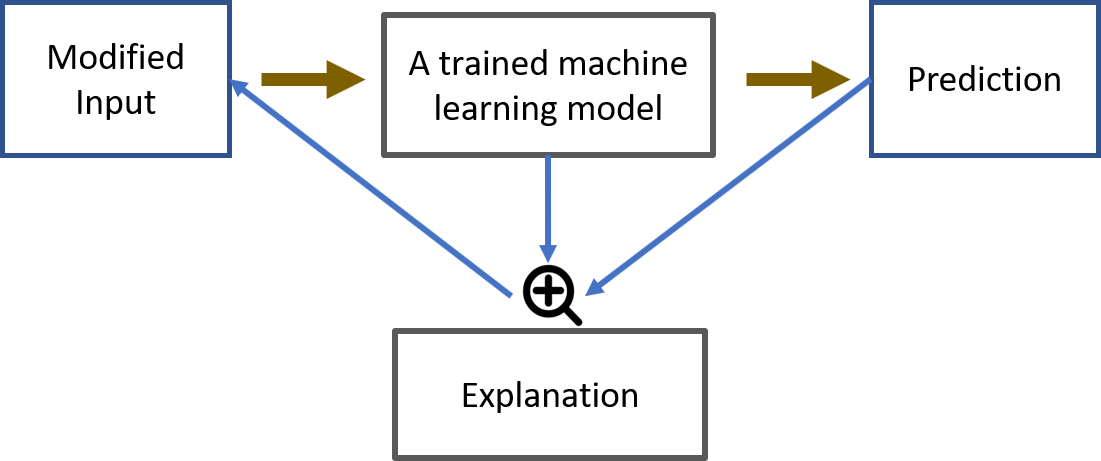}
    \caption{Post-hoc interpretability approach. The explanation method feeds modified input to the trained model, and the model predictions are used along with model internals to reverse engineer the process.}
    \label{fig:post-hoc}
\end{figure}

In terms of complexity, generally, a more complex model is more difficult to interpret and explain~\citep{adadi2018peeking}. The simplest approach for interpretability is to use an intrinsically explainable model that is considered interpretable due to its simple structure, like a decision tree. However, these models usually do not perform as well as the more complex models, which lends credibility to the argument that intrinsic explainability comes with a reduction in prediction performance~\citep{breiman2001statistical}. An alternative approach is \textit{post-hoc interpretability}, which is illustrated in Figure~\ref{fig:post-hoc}. In this approach an explanation method feeds some modified input to a trained machine learning model and uses the predictions and sometimes the model internals to reverse engineer the process. Although this approach might be computationally more expensive, post-hoc interpretability methods tend to be model-agnostic and most recent works in the interpretable AI field falls under this category~\citep{felix2019shap, norgeot2018time, olah2020zoom}.

In terms of scope, we focus on local interpretability methods since they give a more detailed picture of the model behavior. For some models, local explanations are relatively easy to construct. For instance, in a Naive Bayes classification model, we can use the class probabilities with respect to each feature, and for a decision tree, we can use the path chosen as the explanation. However, when there are many features, even these models become increasingly complex and not interpretable. Thus, post-hoc interpretability methods such as LIME~\citep{ribeiro2016should} and SHAP~\citep{lundberg2017unified} can be used to explain decisions of a model.

Another way to clasify interpretable AI methods is based on whether they are model specific or model agnostic. Model agnostic methods are usually preferable when we want to apply the same method to all types of machine learning algorithms. However, model specific methods can use inherent properties of a machine learning algorithm and can be computationally cheaper. A good example for this is the SHAP algorithm proposed by~\citet{lundberg2017unified}. The authors propose kernelSHAP as a model agnostic interpretability method. However, they also propose two faster model-specific approximations of the same approach for neural networks and tree-based models.


Various approaches have been suggested to interpret and understand the behavior of time series classification models. We focus on post-hoc local explanation methods due to their more detailed explanations and ease of use.

Perturbation-based approaches measure the feature importance by replacing the input features with different values and observing the change in the output without any knowledge of model parameters. These methods assign higher feature importance to a feature which has the highest impact on the output when removed. For the time series prediction, representing a removed feature can be tricky, and replacing with mean value and adding random noise are two popular options~\citep{fong2017interpretable}.

Attribution methods explain models by computing the contribution of each input feature to the prediction. Attributions can be assigned using gradient-based methods by measuring the change in the output caused by changes in the input features~\citep{deepliftshrikumar2017learning}. SHAP~\citep{lundberg2017unified} is an attribution based method and has been successfully applied to time series classification~\citep{felix2019shap}.

Evaluation of local explanations remain largely unexplored for time series models. Even though it is studied for computer vision~\citep{bach2015pixel, fong2017interpretable} and natural language~\citep{nguyen2018comparing} these methods cannot be directly applied to time series prediction. In this paper, we propose two new evaluation measures that can be applied to all types of time series forecasting models.

\section{Methodology}
In this section, we describe the datasets, forecasting models and local explanation models used in the experiments. We used the sliding window method for framing the datasets. We applied an 80-20\% train-test split where the last 20\% of the sliding windows are added to the test set.

\subsection{Datasets}
We experiment with two multivariate time series datasets whose characteristics are summarized in Table~\ref{dataset_stats}. We add additional time covariates as features to the original datasets to improve the performance. A min-max normalization is applied on the target feature in both datasets to bound the output feature between 0 and 1. A representative set of 100 randomly selected time series are used to train the machine learning models.
\setlength{\tabcolsep}{6pt}
\renewcommand{\arraystretch}{1.1}
\begin{table}[!htp]
\centering
\caption{Dataset statistics}\label{dataset_stats}
\begin{tabular}{|l|c|c|}
\hline
& \textbf{electricity} & \textbf{Rossmann} \\ \hline
\# time series     & 370         & 1,115 \\ \hline
\# features        & 2           &  7\\ \hline
\# time covariates & 4           &  3 \\ \hline
Sliding window size & 168           &  30 \\ \hline
Granularity     & Hourly      & Daily \\ \hline
\end{tabular} 
\end{table}

\paragraph{Electricity}
This dataset contains the hourly electricity consumption of 370 households in total and has been used as a benchmark for many time series forecasting models~\citep{lim2019temporal, salinas2019deepar}. In addition to two features (hourly electricity consumption and house id), we generate four covariates (hour of the day, day of the week, week of the month, and month) to be included in our analysis.

\paragraph{Rossmann}
Rossmann store sales dataset contains sales data for 1,115 Rossmann stores, providing a useful test bed for sales forecasting tasks. The dataset contains eight features (store number, sales, customers, open, promo, state holiday, school holiday, day of week). We include two additional time covariates, which are week of the month and month.

\subsection{Time Series Models}
We consider three different machine learning models from the literature for time series forecasting: Time Delay Neural Network (TDNN)~\citep{ozyegen2019experimental}, Long Short Term Memory Network (LSTM)~\citep{salinas2019deepar}, and Gradient Boosted Regressor (GBR). There are various strategies for training time series models for time series forecasting~\citep{taieb2012review}. Typically, the multiple input multiple output (MIMO) strategy is used for training the Neural Network models where a single model is trained to predict multiple forecasting horizons. Since we also consider GBR, the MIMO strategy would not be applicable for our analysis. Thus, we use a direct strategy for GBR where a separate model is trained to predict each forecasting horizon.

Both the electricity and the Rossmann datasets contain multiple time series. To maintain generalizability, we randomly selected 100 time series from each dataset and trained the machine learning models on these representative samples.

\subsubsection{Time Delay Neural Network (TDNN)}
This neural network architecture is composed of an input layer, a set of hidden layers and an output layer. The topology of the network architecture is a feedforward neural network. The network is usually described by the number of hidden units at each hidden layer. The architecture we adopted in our analysis is shown in Table~\ref{tab:tdnn_arch}.

\setlength{\tabcolsep}{9pt}
\renewcommand{\arraystretch}{1.1}
\begin{table}[!htp]
    \centering
    \caption{TDNN Network Architecture}\label{tab:tdnn_arch}
    
\begin{tabular}{|c|c|c|}
\hline
\textbf{Layer type} & \textbf{\# of units} & \textbf{Activation} \\ 
\hline
Dense & 64       & Sigmoid    \\
\hline
Dense & 64       & Sigmoid    \\ \hline
\begin{tabular}[c]{@{}c@{}}Dense\\ (Output)\end{tabular} & 12       & - \\ 
\hline
\end{tabular}

\end{table}

The TDNN network used in the experiments contains two hidden layers with 64 hidden units. The sigmoid activation is used after each unit. All of the weights in the model were randomly initialized to be close to $0$. Root Mean Squared Error (RMSE) is selected as our evaluation metric for the loss function and Adam optimizer~\citep{kingma2014adam} is used for backpropagation. 

\subsubsection{Long Short Term Memory Network (LSTM)}
LSTM networks are very popular in the literature for time series analysis. They can capture long term dependencies in sequential data which is an important advantage over TDNNs. LSTMs are a special type of Recurrent Neural Networks; unlike TDNNs they are optimized using backpropagation through time, which unrolls the neural network and backpropagates the error through the entire input sequence. This process can be slow, however it allows the network to take advantage of the temporal dependencies between observations.


Similar to the TDNN model, RMSE is selected as the evaluation metric and Adam optimizer is used for updating the weights of the network. The LSTM architecture we adopted in our analysis is shown in Table~\ref{tab:lstm_arch}.
\setlength{\tabcolsep}{9pt}
\renewcommand{\arraystretch}{1.1}
\begin{table}[!htp]
    \centering
    \caption{LSTM Network Architecture}\label{tab:lstm_arch}
    
\begin{tabular}{|c|c|c|c|}
\hline
\textbf{Layer type} & \textbf{\# units} & \textbf{Activation} & \textbf{Return hidden state} \\ \hline
LSTM & 64       & -          & True \\ 
\hline
LSTM & 64       & ReLU       & False \\ 
\hline
\begin{tabular}[c]{@{}c@{}}Dense\\ (Output)\end{tabular} & 12       & -          & - \\ 
\hline
\end{tabular}
\end{table}

\subsubsection{Gradient Boosted Regressor (GBR)}
Gradient boosting \citep{friedman2001greedy} is another popular model in contemporary machine learning. The simplest gradient boosting algorithm consists of learning many weak learners through functional gradient descent, and adding weak learners to a base function approximator. This is in contrast to how neural networks learn.

Neural networks rose to popularity largely due to their universal function approximator properties \citep{funahashi1989approximate}. Neural networks can be expressed in closed form as
\begin{equation}
\label{eq:hidden}
\mathbf{y} = F(\instanceVect; \neuralNetWeights)
\end{equation}
where $\neuralNetWeights$ represents the weights in the network, $\instanceVect$ is a vector consisting of the regressor values, and $\mathbf{y}$ is the target vector. Neural networks utilize activation functions to become a non-linear function approximator. Gradient boosting utilizes classification and regression trees (CART) \citep{breiman1984classification} instead of activation functions to achieve its non-linearity. That is,
\begin{equation}
\label{eq:grad_boosting}
\mathbf{y} = F(\mathbf{x}; \{\beta_{m}, \mathbf{a}_m\}_{m=1}^M)=\sum_{m=1}^{M} \beta_{m} h\left(\mathbf{x} ; \mathbf{a}_{m}\right)
\end{equation}
where $h\left(\mathbf{x} ; \mathbf{a}_{m}\right)$ represent the weak learners. In addition, the number of learners is specified by a hyper-parameter, $M$. By changing the atomic building block from neurons and activation functions to decision trees, several modifications are needed to be made. Instead of performing classical gradient descent, more complicated updates are required for training. These updates are made through performing functional gradient descent, and adding these functions to the base model. 

In equation \eqref{eq:grad_boosting}, the parameters $\mathbf{a}_m$ represent the weights inside each decision tree and the parameter $\beta_m$ represents the weight of the tree in the general model. The loss function can be chosen according to the problem specifications, where commonly used loss functions include the mean squared loss ($L2$), mean absolute loss ($L1$) and logistic loss.


In our analysis, due to the multi-step forecast nature of the problem, we have trained numerous gradient boosting regressors, each responsible for predicting a particular forecasting horizon. Each submodel contains the same hyper parameters, listed in Table \ref{tab:gbr_params}.

\setlength{\tabcolsep}{9pt}
\renewcommand{\arraystretch}{1.1}
\begin{table}[!htp]
    \centering
    \caption{Gradient Boosting Regressor Parameters}\label{tab:gbr_params}
    \begin{tabular}{|l|c|}
\hline
\textbf{Loss function} & Least squares regression \\
\hline
\textbf{Learning rate} & 0.01 \\
\hline
\textbf{\# of trees} & 100 \\
\hline
\textbf{Tree splitting criterion} & Friedman Mean Squared Error \\
\hline
\textbf{Max depth} & 3 \\
\hline
\end{tabular}
\end{table}

\subsection{Local Explanation Methods}
A particular dataset can be defined as combination of $\numFeatsTotal$ features.  Each feature $j \in \{1,\hdots,\numFeatsTotal\}$ has a corresponding feature space, which we denote as $\setFeature^\indexFeat$, with $n$ permissible values. That is,
\begin{align*}
    \setFeature^j \equiv \{\setFeatureVal_1^{[\indexFeat]}, \setFeatureVal_2^{[\indexFeat]}, \hdots, \setFeatureVal_n^{[\indexFeat]}\} = \{\setFeatureVal_i^{[\indexFeat]}\}_{i=1}^{n}
\end{align*}

A discrete time series model, $\mathbf{F}$, can be formulated as $\mathbf{y}_t = \mathbf{F}(\instanceMat_t) + \boldsymbol{\epsilon}_t$, where $t$ represents the time step. The explanatory variables are represented as $\instanceMat_t$, the target vector is $\mathbf{y}_t$, and the error in the model is $\boldsymbol{\epsilon}_t$. The target vector, $\mathbf{y}_t$ produces a multi-horizon forecast of length $\timeHorizonTotal$. For simplicity of notation, we only consider one of these target time horizons, at an arbitrary index $\timeHorizon \in \{1,\hdots,\timeHorizonTotal\}$, in below analytical expressions.
\begin{align*}
    y_t &\equiv \mathbf{y_t}^{[\timeHorizon]} \\
    &= \mathbf{F}_{\timeHorizon}(\instanceMat_t) + \boldsymbol{\epsilon}_t^{[\timeHorizon]} \\
    &\equiv f(\instanceMat_t) + \epsilon_t
\end{align*}
We will drop the index, $\timeHorizon$, without loss of generality, and refer to this by scalar notation $y_t$. It is a simple extension to perform a multi-step forecast. As well, we explicitly chose to omit the time series index in order to increase legibility. Note that our proposed approach naturally scales to multiple time series.

We denote the explanatory variables in matrix form, to allow for a sliding window of time slices, $\instanceVect_{t}$. The sliding window is of length $\windowSize$. This can be seen as,
\begin{align*}
\instanceMat_t &= \begin{bmatrix}
      \instanceVect_{t-(\windowSize-1)}, & \instanceVect_{t-(\windowSize-2)}, & \hdots, & \instanceVect_{t}
\end{bmatrix} \\
&= \{ x_{\indexFeat\indexSlice,t} \}, \quad \indexFeat \in \{1,\hdots,\numFeatsTotal\},\quad \indexSlice \in \{1,\hdots,\windowSize\}
\end{align*}
where we describe an individual covariate at position $(\indexFeat,\indexSlice)$ in the full covariate matrix at time step $t$ as $x_{\indexFeat \indexSlice,t}$. 

We take $\instanceVect_t$ as an individual time slice. Each feature relates to its feature set, represented by a superscript. That is,
\begin{align*}
\instanceVect_t &= \begin{bmatrix}
    \instance_t^{[1]} \\
    \instance_t^{[2]} \\
    \vdots \\
    \instance_t^{[\numFeatsTotal]}
\end{bmatrix}
\end{align*}

\noindent For example, if our sets were
\begin{align*}
\setFeature^1 &\equiv \{\text{Apple}, \text{Banana}, \text{Tomato}\} \\
\setFeature^2 &\equiv \{\text{Red}, \text{Yellow}, \text{Green}\} \\
\setFeature^3 &\equiv \{\text{Ripe}, \text{Not Ripe}\} \\
\setFeature^4 &\equiv \{\text{Bruised}, \text{Not Bruised}\}
\end{align*}

\noindent then we can represent a ripe tomato at time $t$ as:
\begin{align*}
\instanceVect_t &= \begin{bmatrix}
      \text{Tomato} \\
      \text{Red} \\
      \text{Ripe} \\
      \text{Not Bruised}
     \end{bmatrix}
\end{align*}

\noindent and a sample sliding window, $\instanceMat_t$, for $\windowSize=3$, can be as follows:
\begin{align*}
\instanceMat_t &= \begin{bmatrix}
  \text{Tomato} & \text{Tomato} & \text{Tomato} \\
  \text{Green} & \text{Red} & \text{Red} \\
  \text{Not Ripe} & \text{Not Ripe} & \text{Ripe} \\
  \text{Not Bruised} & \text{Not Bruised} & \text{Not Bruised}
 \end{bmatrix}
\end{align*}

In post-hoc local explanation methods, we can give an importance matrix $\impScoreMat_t = \{\impScore_{\indexFeat \indexSlice,t}\}$ to provide feature importance scores for an instance $\instanceMat_t = \{x_{\indexFeat \indexSlice,t}\}$ at a given time $t$. A local explanation method aims to find the importance of each covariate. 

We consider two local explanation methods, namely, omission and SHAP, and compare the performances against a baseline.
\subsubsection{Random Baseline}
In this approach, we randomly rank features from most important to least important. 

\subsubsection{Omission}
In omission, the estimated importance of a regressor in question, $\instance_{\indexFeat \indexSlice,t}$, is denoted as $\impScore_{\indexFeat \indexSlice,t}$. This importance score is found by removing the regressor from the sliding window matrix, which we represent as $\instanceMat_{t \backslash \instance_{\indexFeat \indexSlice,t}}$, and measuring the effect. That is, 
\begin{equation}
\impScore_{\indexFeat \indexSlice,t} = f(\instanceMat_t) - f(\instanceMat_{t \backslash \instance_{\indexFeat \indexSlice,t}})
\end{equation}

Unlike a natural language processing problem where we can easily remove words by replacing it with zeros, we cannot just remove a feature with zeros without consequences in the regression setting. If we replace features with zeros, then the interpretability method will learn that when a covariate $x_{\indexFeat \indexSlice,t}$ is zero, it has no contribution to the prediction~\citep{sturmfels2020visualizing}. Alternative approaches can be replacing removed features with local mean, global mean, local noise and global noise. Note that local refers to a single sample and global refers to all the samples of that feature in the dataset. On the other hand, adding local and global noise can put extra peaks and slopes to the input, which are usually important for the prediction. Thus, we choose to test the omission method with local and global mean replacement. The local mean calculates the average value of a feature in a given window slice. This is done by performing a column-wise average over a window in $\instanceMat_t$. The global mean is time invariant, and is the average feature value of a given time series of length $\timeVarTotal$.
\begin{align}
    \text{Local Mean: } & \mu^{\text{loc}}_{\indexFeat,t} = \dfrac{1}{\windowSize} \sum_{\indexSlice=1}^{\windowSize} x_{\indexFeat \indexSlice,t}\\
    \text{Global Mean: } & \mu_\indexFeat^{\text{glo}} = \dfrac{1}{\timeVarTotal} \sum_{t=1}^{\timeVarTotal} \instance^{[\indexFeat]}_{t} 
\end{align}


\subsubsection{SHAP}
SHAP is a post-hoc and model agnostic approach which follows a very similar logic to many other popular interpretability methods such as LIME~\citep{ribeiro2016should} and DeepLIFT~\citep{deepliftshrikumar2017learning}. All these methods follow the same core logic where they learn a local linear model to explain a more complex model. As such, these methods are also referred to as additive feature attribution methods. SHAP is the only local explanation approach that satisfies three desirable properties: local accuracy, missingness, and consistency~\citep{lundberg2017unified}.

In our analysis, we use two SHAP based approaches, DeepSHAP~\citep{lundberg2017unified} and TreeExplainer~\citep{lundberg2020local}, and two model-specific methods that approximate SHAP values for neural networks and tree-based models, respectively. We experiment with DeepSHAP method for TDNN and LSTM models, and TreeExplainer for the GBR models, and use the shap library in Python \citep{lundberg2017unified}.

\subsection{Evaluation Metrics}
We propose two new evaluation metrics for local explanation methods in time series forecasting models. 

We can organize the top $K$ features according to  a local explanation metric. These are sorted according to largest $\impScore_{ij,t}$ values. When we take out the top $\numFeatsRemoved$ features from $\instanceMat_t$, which, without loss of generality, is defined as $\instanceMat_{t,\backslash {1:\numFeatsRemoved}}$. This is a combination of defined covariates, $x_{\indexFeat \indexSlice,t}$ and \textit{random covariates}, $r_{\indexFeat \indexSlice,t}$, sampled from the marginal distribution of the respective feature space $\setFeature^j$.

For example, a particular model may have a window of length $\windowSize=2$ with three features at each time slice, $\numFeatsTotal=3$. Then, a local explanation model determines the importance of variables $\impScoreMat_t$, where $\impScore_{12,t} > \impScore_{31,t} > \impScore_{\indexFeat \indexSlice,t} \quad \forall (\indexFeat,\indexSlice) \notin \{(1,2),(3,1)\}$. We represent the removal of the top two most important features (i.e. $(\indexFeat=1, \indexSlice=2)$ and $(\indexFeat=3, \indexSlice=1)$) as
\begin{align*}
\instanceMat_{t,\backslash 1:2} &= \begin{bmatrix}
      x_{11,t} & \tcolB{r_{12,t}} \\
      x_{21,t} & x_{22,t} \\
      \tcolB{r_{31,t}} & x_{32,t} \\
\end{bmatrix} 
\end{align*}

Due to the stochastic nature of this process, in order to perform the feature ablation, we need to collect the expected value of the ablated feature. We represent this as $\expectedVal[\cdot]$.

We next define two evaluation metrics to evaluate the \textit{local fidelity} of local explanation methods. Local fidelity is an important measure for explanation methods. It evaluates the level of alignment between the interpretable model and the black-box model~\citep{guidotti2018survey}. Local states that we are looking for this alignment is the neighborhood of an instance.

Local fidelity can be measured by $\numFeatsRemovedInc$-ablation methods~\citep{arras2016explaining, sturmfels2020visualizing}, where we delete features in the order of their estimated importance for the prediction. \citet{nguyen2018comparing} uses two metrics to measure local fidelity: Area Over the Perturbation Curve (AOPC) and Switching Point (SP). However, similar to other existing evaluation methods, AOPC and SP are only used for classification tasks.

To measure local fidelity for a multivariate time series forecasting task, we define two new metrics: \textit{AOPCR} and \textit{APT}. These metrics are slight variants of AOPC and SP, designed for evaluating interpretable AI methods for time series forecasting task.


AOPCR and APT measure the local fidelity in two different ways. More specifically, AOPCR measures the effect of removing the top $\numFeatsRemoved$ features and APT measures the percentage of features that need to be removed to pass a certain threshold. AOPCR focuses on a small percentage of the most important features whereas APT usually requires the removal of a higher percentage of features. 

\subsubsection{Area Over the Perturbation Curve for Regression (AOPCR)}
The area over the perturbation curve for regression at time horizon $\timeHorizon$, denoted as $\text{AOPCR}_\timeHorizon$, is obtained as
\begin{equation}
\label{eq:hidden}
\text{AOPCR}_\timeHorizon = \frac{1}{\numFeatsRemoved} \sum_{\numFeatsRemovedInc=1}^{\numFeatsRemoved}\mathbf{F}_\timeHorizon(\instanceMat_t) - \mathbf{F}_{\timeHorizon}(\instanceMat_{t,\backslash 1:\numFeatsRemovedInc})
\end{equation}
Then, the total area over the perturbation for regression is the average of all the time steps $\timeHorizon = 1,\hdots,\timeHorizonTotal$, where
\begin{equation}
\label{eq:hidden}
\text{AOPCR} = \frac{1}{\timeHorizonTotal}\sum_{\timeHorizon=1}^{\timeHorizonTotal}\text{AOPCR}_
\timeHorizon
\end{equation}
In its current state, AOPCR introduced random variables due to the feature removal procedure. In order to explicitly calculate AOPCR and AOPCR$_t$, we collect the expected value of the ablated features, and compute AOPCR$_\timeHorizon$ with this expected value, denoted by $\widehat{\text{AOPCR}}_\timeHorizon$. That is,
\begin{align}
\widehat{\text{AOPCR}_\timeHorizon} &= \expectedVal[\text{AOPCR}_\timeHorizon] \\
&= \frac{1}{\numFeatsRemoved} \sum_{\numFeatsRemovedInc=1}^{\numFeatsRemoved}F_\timeHorizon(\instanceMat_t) - \expectedVal\left[F_\timeHorizon(\instanceMat_{t,\backslash 1:\numFeatsRemovedInc})\right]
\end{align}

\noindent where the source of randomness, $r$, is the randomly drawn covariates, $r_{\indexFeat \indexSlice,t}$. Similarly, $\widehat{\text{AOPCR}} = \expectedVal[\text{AOPCR}]$.

\smallskip
\subsubsection{Ablation Percentage Threshold (APT)}
APT provides an alternative way to measure local fidelity. In classification, the switching point~\citep{nguyen2018comparing} is defined as the percentage of features that need to be deleted before the prediction switches to another class. For regression, we can define the switching point as a point above and below the original prediction by a predefined threshold distance.

In this approach, we take all $\numFeatsTotal$ features and sort them by importance. Then, we remove $\numFeatsRemoved$ features from the top or the bottom, stopping when the prediction changes by a pre-defined factor, $\aptThreshold$. The percentage of features that need to be removed until the prediction passes the threshold is reported as the APT score at the particular time step. A lower APT score means a lower percentage of features had to be deleted to pass the threshold, which shows a higher local fidelity. 

We can define APT at time horizon $\timeHorizon$ with significance factor $\aptThreshold$ as follows.
\begin{align}
\label{eq:hidden}
\text{APT}_{\timeHorizon,\aptThreshold}\  &= \argmin_{\numFeatsRemovedInc \in \{1,\hdots,\numFeatsTotal\}}\frac{\numFeatsRemovedInc}{\numFeatsTotal} \\
\text{such that:} \ & \quad \mathbf{F}_{\timeHorizon}(\instanceMat_t) (1 + \aptThreshold) >  \mathbf{F}_{\timeHorizon}(\instanceMat_{t,\backslash 1:\numFeatsRemovedInc})
\end{align}

\noindent Note that in order find the lower bound significance threshold, $\aptThreshold$ needs to be set to a negative number. This represents when a predicted value has gotten significantly smaller due to feature removal. The total ablation percentage threshold is a simple average over the time index, that is,
\begin{equation}
\label{eq:hidden}
\text{APT}_{\aptThreshold} = \frac{1}{\timeHorizonTotal}\sum_{\timeHorizon=1}^{\timeHorizonTotal}\text{APT}_{\timeHorizon,\aptThreshold}
\end{equation}
Finally, to convert the theoretical metric, APT$_\aptThreshold$, into an experimental metric, we take the expected value, $\widehat{\text{APT}}_\aptThreshold = \expectedVal[\text{APT}_\aptThreshold]$

\subsubsection{Implementation Details}
We estimated the expected values of the metrics by taking numerous Monte Carlo samples. The ablated features were randomly replaced with in sample values, proportional to the ablated feature's distribution.

Any biasing choice in the design of the experiment can be a threat to the external validity. To evaluate the explanation methods, we arbitrarily choose the $\numFeatsRemoved$ value and a threshold value for AOPCR and APT methods, respectively. A very small value can make the AOPCR and APT methods too sensitive and make the resulting scores incomparable. Thus we found a $\numFeatsRemoved$ value of 10 for AOPCR to be suitable for the analysis. Additionally for APT, a very large value can make it impossible for an ablated sample to pass the threshold and again result in incomparable scores. For the threshold value in APT, we experimented with multiple thresholds and found that 10\% is a good value for comparing different methods and models for both datasets.

In order to reduce the randomness in the evaluation metrics, we added an early stopping condition. Once the margin of error for the $\widehat{\text{AOPCR}}$ (or $\widehat{\text{APT}}$) statistic was less than 0.05\%, with 95\% confidence (assuming a Gaussian distribution of the statistic), we ceased taking more samples. This was a natural stopping condition, which provided stable results.

A summary statistic can often contain a threat to the conclusion validity. In evaluations, we need to take the average of the scores over Monte Carlo samples and forecasting horizons, $t_0$. However, since the important features for the model can vary over $\tau \in \{1,\hdots,t_0\}$, the scores can also vary across in the same interval. Therefore, we compute the confidence intervals separately for each forecasting horizon. Once every forecasting horizon lies in a tolerable confidence interval, the scores are first averaged over Monte Carlo samples and then over $t_0$.

\section{Results}

\subsection{Model Performances}
The performance of the three models (LSTM, TDNN and GBR) are compared on the Electricity and the Rossmann dataset. The Normalized Root Mean Squared Error (NRMSE) and Normalized Deviation (ND) scores on the test set is presented in Table~\ref{tab:model_perf}. Explicit expressions for these performance metrics are provided below.
\begin{align}
    \text{NRMSE}(y, \hat{y}) = \frac{\displaystyle \sqrt{\frac{1}{N} \sum_{i=1}^{N}(\hat{y_i} - y_i)^2}}{\displaystyle y_{max}-y_{min}}, \qquad
    \text{ND}(y, \hat{y}) = \frac{\displaystyle \sum_{i=1}^{N}\vert\hat{y_i} - y_i\vert}{\displaystyle \sum_{i=1}^{N} \vert y_i \vert}
\end{align}

\begin{table}[hbt!]
\centering
\caption{Comparison of GBR, LSTM and TDNN models on Electricity and Rossmann datasets. GBR performs the best for both datasets, and TDNN performs the worst.}
\label{tab:model_perf}
\begin{tabular}{|c|c|c|c|c|}
\hline
\multicolumn{1}{|l|}{}               & \multicolumn{2}{c|}{\textbf{Electricity}} & \multicolumn{2}{c|}{\textbf{Rossmann}} \\ \hline
\multicolumn{1}{|l|}{\textbf{Model}} & \textbf{NRMSE}      & \textbf{ND}         & \textbf{NRMSE}     & \textbf{ND}       \\ \hline
TDNN                                 & 0.160               & 0.294               & 0.190              & 0.393             \\ \hline
LSTM                                 & 0.081               & 0.141               & 0.139              & 0.260             \\ \hline
GBR                                  & \textbf{0.046}      & \textbf{0.072}      & \textbf{0.128}     & \textbf{0.211}    \\ \hline
\end{tabular}
\end{table}

\begin{figure}[hbt!]
    \centering
        \subfloat[Electricity Dataset ]{\includegraphics[width=0.5\textwidth]{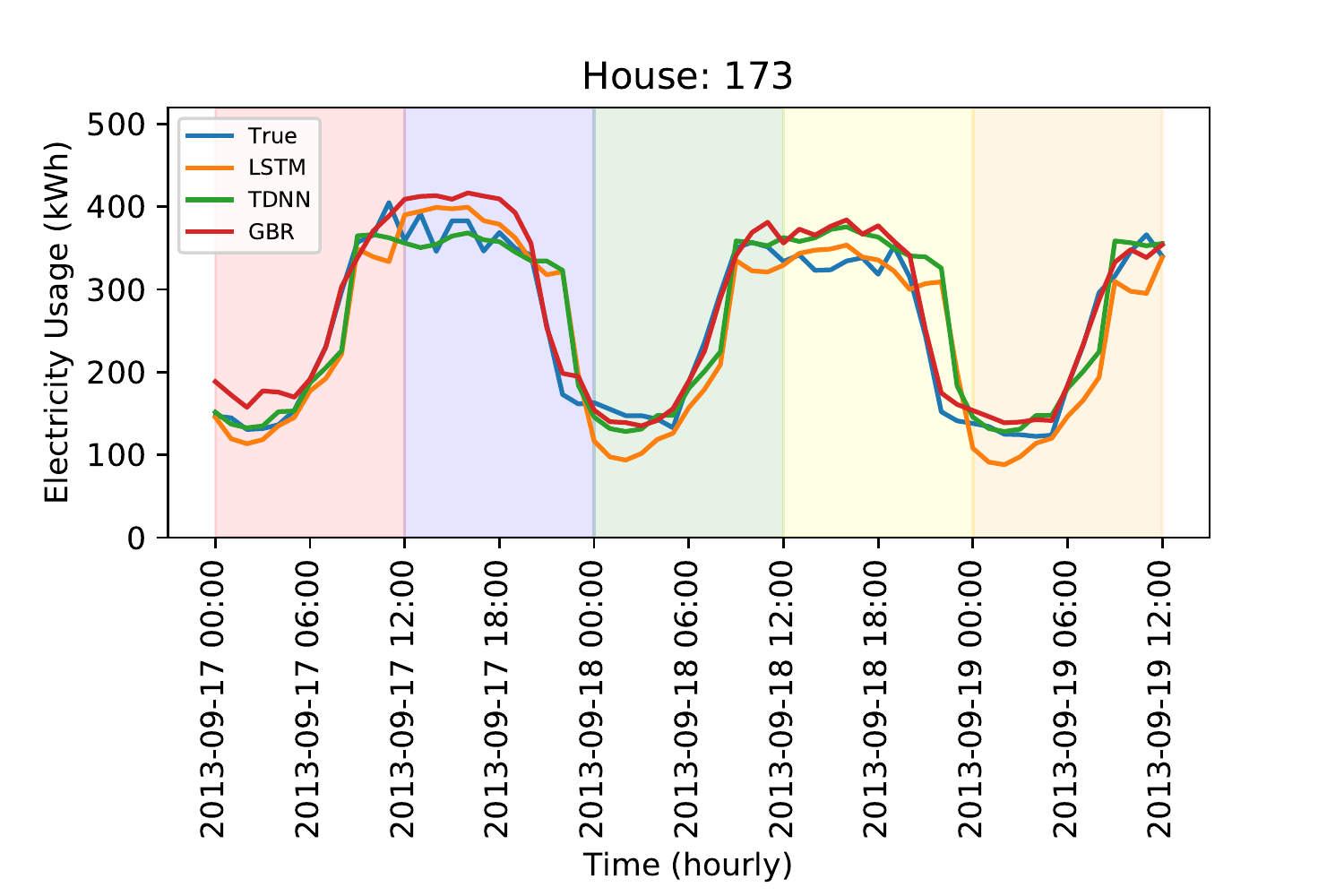}
        \label{fig:elect_pred_vis}
        }
        \subfloat[Rossmann Dataset ]{\includegraphics[width=0.5\textwidth]{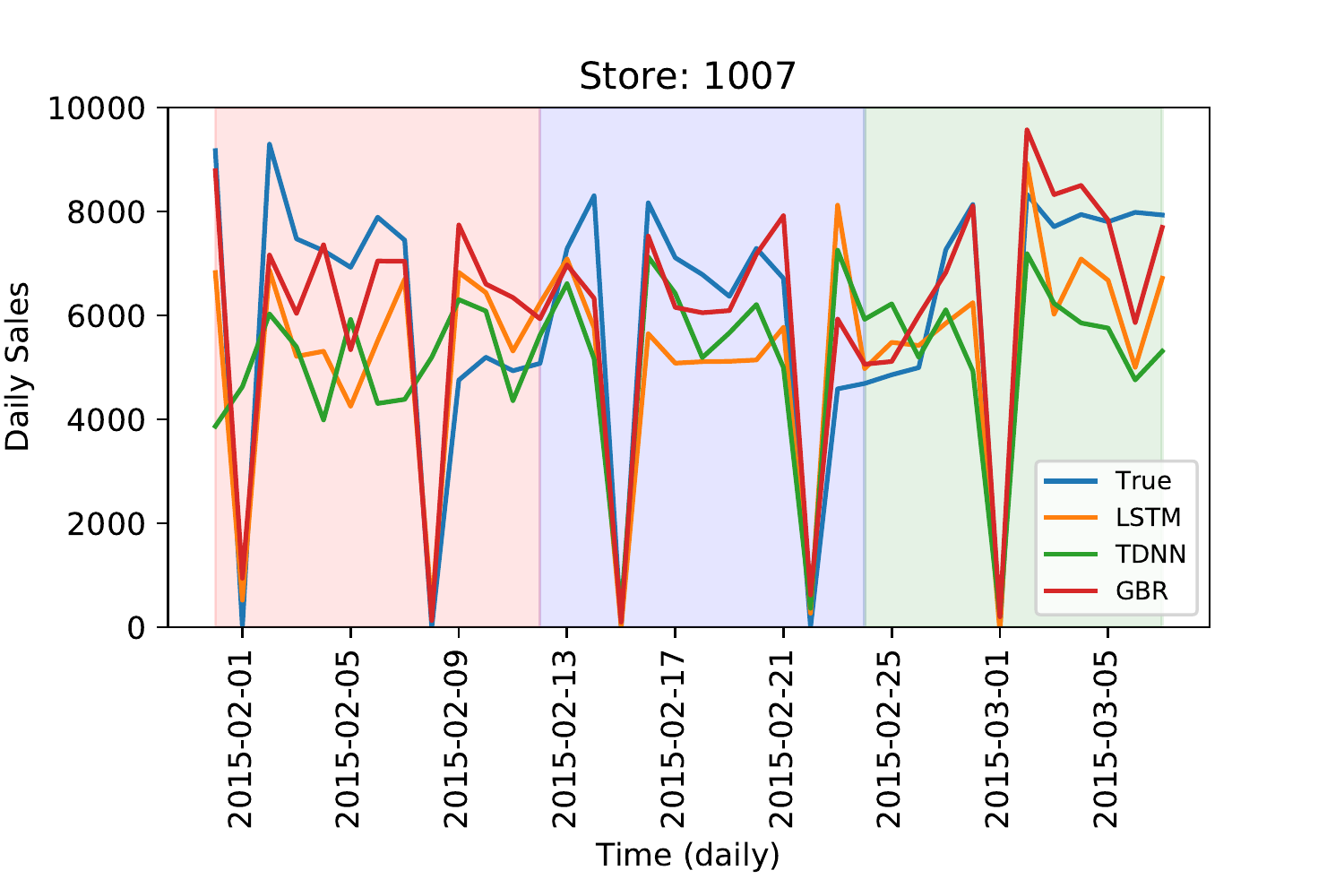}
        \label{fig:ross_pred_vis}
        }\\
    \caption{Visualization of models' predictions on a random series. Each background color corresponds to a separate prediction window. Each model generates predictions that can capture the trends for the provided sample.}
    \label{fig:visualize_pred}

    \end{figure}

Overall the GBR model performed best and the TDNN model performed significantly worse than the other models for both metrics. The analysis of the individual samples show that the TDNN model is able to capture some of the temporal behavior, although not as much as the other models. Thus, the TDNN model is kept as a simple baseline and it is used for evaluating the local explanation methods. Also note that, the clear performance ranking of the three prediction models (i.e. TDNN $\prec$ LSTM $\prec$ GBR) is useful for understanding the link between the model performance and the interpretability.  To visually illustrate the models' behavior, we present visualizations of the models' predictions on a randomly selected test sample in Figure~\ref{fig:visualize_pred}. Each background color on the figures correspond to a prediction window with a size of 12 timesteps.

\subsection{Evaluation of local explanations}
The comparison of AOPCR and APT scores for the electricity and Rossmann datasets are shown in Table~\ref{tab:aopcr_All_results} and Table~\ref{tab:apt_All_results}, respectively.



\setlength{\tabcolsep}{9pt} 
\renewcommand{\arraystretch}{1.2} 
\begin{table*}[!htp]
\centering
\caption{AOPCR scores for electricity and Rossmann datasets. Lower percentage shows higher local fidelity. Best method in each column is in bold.}
\label{tab:aopcr_All_results}
    \subfloat[electricity dataset \label{tab:aopcr_electricity_results}]{
    \scalebox{0.9}{
    \begin{tabular}{|l|r|r|r|r|r|r|}
\hline
 & \multicolumn{2}{c|}{TDNN} & \multicolumn{2}{c|}{LSTM} & \multicolumn{2}{c|}{GBR} \\ \hline
Model & positive & negative & positive & negative & positive & negative \\ \hline
Random & -0.00050 & -0.00035 & 0.00000 & -0.00014 & 0.00008 & -0.00003 \\ \hline
Omission (Global) & \textbf{0.00263} & \textbf{-0.00573} & 0.08568 & -0.13806 & 0.06307 & -0.05938 \\ \hline
Omission (Local) & 0.00243 & -0.00527 & 0.06884 & -0.07899 & 0.04892 & -0.04395 \\ \hline
SHAP & 0.00055 & -0.00430 & \textbf{0.11379} & \textbf{-0.14165} & \textbf{0.07067} & \textbf{-0.06664} \\ \hline
\end{tabular}}} \\
    \subfloat[Rossmann dataset \label{tab:aopcr_rossmann_results}]{
    \scalebox{0.9}{
    \begin{tabular}{|l|l|l|l|l|l|l|}
\hline
 & \multicolumn{2}{c|}{TDNN} & \multicolumn{2}{c|}{LSTM} & \multicolumn{2}{c|}{GBR} \\ \hline
Model & positive & negative & positive & negative & positive & negative \\ \hline
Random & 0.00011 & -0.00014 & 0.00003 & -0.00019 & -0.00090 & -0.00129 \\ \hline
Omission (Global) & 0.00344 & -0.00279 & 0.08478 & -0.09673 & 0.06600 & -0.09787 \\ \hline
Omission (Local) & 0.00150 & -0.00126 & 0.07034 & -0.08011 & 0.06071 & -0.09026 \\ \hline
SHAP & \textbf{0.03734} & \textbf{-0.02971} & \textbf{0.09522} & \textbf{-0.09802} & \textbf{0.07862} & \textbf{-0.11273} \\ \hline
\end{tabular}}}
\end{table*} 

\setlength{\tabcolsep}{9pt} 
\renewcommand{\arraystretch}{1.2} 
\begin{table*}[!htp]
\centering
\caption{APT \% scores for electricity and Rossmann datasets. Lower percentage shows higher local fidelity. Best method in each column is in bold.}
\label{tab:apt_All_results}
    \subfloat[electricity dataset \label{tab:apt_electricity_results}]{
    \scalebox{0.9}{
    \begin{tabular}{|l|r|r|r|r|r|r|}
\hline
 & \multicolumn{2}{c|}{TDNN} & \multicolumn{2}{c|}{LSTM} & \multicolumn{2}{c|}{GBR} \\ \hline
Model & positive & negative & positive & negative & positive & negative \\ \hline
Random & 0.991 & 0.800 & 0.523 & 0.508 & 0.725 & 0.655 \\ \hline
Omission (Global) & 0.883 & 0.682 & \textbf{0.130} & \textbf{0.114} & 0.477 & 0.493 \\ \hline
Omission (Local) & 0.900 & 0.707 & 0.132 & 0.177 & 0.524 & 0.532 \\ \hline
SHAP & \textbf{0.642} & \textbf{0.363} & 0.141 & 0.130 & \textbf{0.348} & \textbf{0.423} \\ \hline
\end{tabular}}} \\
    \subfloat[Rossmann dataset \label{tab:apt_rossmann_results}]{
    \scalebox{0.9}{
    \begin{tabular}{|l|r|r|r|r|r|r|}
\hline
 & \multicolumn{2}{c|}{TDNN} & \multicolumn{2}{c|}{LSTM} & \multicolumn{2}{c|}{GBR} \\ \hline
Model & positive & negative & positive & negative & positive & negative \\ \hline
Random & 0.842 & 0.908 & 0.362 & 0.460 & 0.611 & 0.600 \\ \hline
Omission (Global) & 0.810 & 0.890 & \textbf{0.093} & \textbf{0.204} & 0.232 & 0.369 \\ \hline
Omission (Local) & 0.813 & 0.893 & 0.160 & 0.273 & 0.263 & 0.393 \\ \hline
SHAP & \textbf{0.631} & \textbf{0.726} & 0.117 & 0.248 & \textbf{0.150} & \textbf{0.226} \\ \hline
\end{tabular}}}
\end{table*}

First, we compare the explanation methods for each model. For the GBR model, SHAP performed significantly better in both datasets for both metrics. Thus, results suggest that SHAP works well for explaining tree-based models. SHAP is easily able to identify the important features in the GBR model.

Interestingly, in the experiments with the LSTM model, SHAP method produced better AOPCR scores but worse APT scores compared to global omission. Also considering its computational burden, we recognize that SHAP method is not preferable to use, and the global omission is preferred due to its simplicity without loss of accuracy.

For the TDNN models, we observed that the omission methods fail to identify important features. This is observed because the produced AOPCR and APT scores are very similar to random feature removal. For instance for the Rossmann dataset, there is less than 4\% difference in APT scores between the Random and Omission methods. This suggests that omission explanation methods fail to explain the TDNN model in question and not be used as a local explanability method here. On the other hand, SHAP was still able to capture the important features for the model even though the model performance was low.



Secondly, we can compare the explanation methods independent of the machine learning model used. In that case, random explanations led to the worst scores for all cases as expected. The APT scores can be high for random explanations, even close to 100\%. This happens because for a given sample, if all the features are removed and the prediction did not pass the threshold, the APT scores of 100\% is assigned. 

Global omission method outperformed the local omission in all cases for both datasets which indicates global omission is preferable over local omission as a local explanation method for time series. Overall, SHAP method was able to, in general, best explain all models followed by the global omission method. This is not surprising since SHAP method is more complex and it also considers the interactions between the features unlike the Omission methods. However, results indicate that it can perform equally or worse than simpler approaches in select examples, implying that many explanation models should be tested in time series forecasting experiments.

Finally, we compare the evaluation methods. AOPCR and APT give two different views on local fidelity. Our experiments show these local fidelity scores do not have to correlate, therefore each method should be used according to the intended experiment. If the experiment cares more about correctly identifying the importance of a predetermined number of top features, AOPCR should be used. Based on this, SHAP method is preferable for the LSTM and the GBR models whereas the results are indeterminate for the TDNN model. On the other hand, if the experiment instead cares about the general explainability of the model, APT should be used. In this case, SHAP is preferable for the TDNN and the GBR model whereas global omission is preferable for the LSTM model.


\section{Conclusion and Future Work}
There has been a significant interest in AI interpretability caused by a growing adoption of machine learning systems. Even though there are some studies focusing on the interpretability of the machine learning models for time series, most of the existing literature is focused on the time series classification task. There is relatively low research interest in interpreting time series forecasting models. An improved understanding on evaluating local explanations can contribute to a further progress in the area. Thus, we focus on evaluating local explanation methods for multivariate time series forecasting problem.

Local explanations are typically computed by finding the importance of features towards the prediction. In this study, two new evaluation measures are proposed for thorough comparison of the local explanation methods. Three local explanation methods are compared for multivariate time series forecasting problem. More specifically, we first trained three models (TDNN, LSTM, GBR) on two datasets (Electricity and Rossmann). Then, we evaluated the three local explanation methods for all the models using two new local fidelity measures suitable for time series forecasting tasks. Overall, we found that SHAP method has the highest fidelity, especially for tree-based models, and global mean replacement is a preferable choice over local mean replacement for both datasets. Additionally, we showed the evaluation scores can vary across different datasets and models.


An area that could be further explored is the idea of placing more weight on immediate time steps. That is, we can modify the evaluation metrics as 
\begin{align*}
\label{eq:hidden}
\text{AOPCR} &= \frac{1}{\timeHorizonTotal}\sum_{\timeHorizon=1}^{\timeHorizonTotal}\gamma^{\timeHorizon-1}\text{AOPCR}_\timeHorizon \\
\text{APT}_{\alpha} &= \frac{1}{\timeHorizonTotal}\sum_{\timeHorizon=1}^{\timeHorizonTotal}\gamma^{\timeHorizon-1}\text{APT}_{\timeHorizon,\alpha}
\end{align*}
By setting $\gamma = 1$, these metrics can be reduced back to our initially proposed evaluation metrics. As $\gamma \rightarrow 0$, more weight is placed on the initial terms in the expansion. The choice of $\gamma$ could be application specific. For example, if a short-term weather model is being evaluated, it might be more important to predict the next day's forecast compared to forecasting five days into the future. 

\section{Data Availability Statement}
The data that supports the findings of this study are openly available in UCI Machine Learning and Kaggle repositories at: \\ 
\url{https://www.kaggle.com/c/rossmann-store-sales/data} \\
\url{https://archive.ics.uci.edu/ml/datasets/Individual+household+electric+power+consumption}

\hfill \break
{\bf Acknowledgement.} The authors would like to thank Merve Bodur for valuable discussions throughout this work. The authors also would like to thank LG Sciencepark for supporting this project.

\bibliography{refs}
\end{document}